\documentclass[letterpaper, 10 pt, conference]{ieeeconf}  

\IEEEoverridecommandlockouts                              

\usepackage{graphics} 
\usepackage{epsfig} 
\usepackage{times} 
\usepackage{amsmath} 
\usepackage{amssymb}  
\usepackage{cite}
\usepackage{amsmath,amssymb,amsfonts}
\usepackage{algorithmic}
\usepackage{graphicx}
\usepackage{textcomp}
\usepackage{xcolor}
\usepackage{lipsum}
\usepackage{array}
\usepackage{booktabs}
\usepackage{tabularx}
\usepackage{multicol}
\usepackage{multirow}
\usepackage{xcolor}
\usepackage{amssymb}
\usepackage{amsmath}
\usepackage{stfloats}
\usepackage{mathdots}
\usepackage{graphicx} 
\usepackage{caption}
\usepackage{subfigure}
\usepackage{algorithm}
\usepackage{bm}
\usepackage{listings}
\usepackage{hyperref}
\usepackage{caption}
\usepackage{subcaption}
\usepackage{svg}

\usepackage{xcolor}

\definecolor{codegreen}{rgb}{0,0.6,0}
\definecolor{codegray}{rgb}{0.5,0.5,0.5}
\definecolor{codepurple}{rgb}{0.58,0,0.82}
\definecolor{backcolour}{rgb}{0.95,0.95,0.92}

\lstdefinestyle{mystyle}{
    backgroundcolor=\color{backcolour},   
    commentstyle=\color{codegreen},
    keywordstyle=\color{magenta},
    numberstyle=\tiny\color{codegray},
    stringstyle=\color{codepurple},
    basicstyle=\ttfamily\footnotesize,
    breakatwhitespace=false,         
    breaklines=true,                 
    captionpos=b,                    
    keepspaces=true,                 
    numbers=left,                    
    numbersep=5pt,                  
    showspaces=false,                
    showstringspaces=false,
    showtabs=false,                  
    tabsize=2
}
\DeclareUnicodeCharacter{FF0C}{,}
\title{\LARGE \bf
LLM A*: Human in the Loop Large Language Models Enabled A* Search for Robotics
}

\author{Hengjia Xiao$^{1}$, Peng Wang$^{2}$, Mingzhe Yu$^{3}$, Mattia Robbiani$^{2}$
\thanks{*This work was supported by Manchester Metropolitan University}
\thanks{Peng Wang and Mattia Robbiani are with the Department of Computing and Mathematics, Manchester Metropolitan  University, Manchester, M15 6BH, UK. Hengjia Xiao is a Liverpool University graduate who participates in the work as an independent researcher. Mingzhe Yu is with Johnson Matthey Technology Centre, Johnson Matthey.
        {\tt\small p.wang@mmu.ac.uk}}%
}

\begin{document}

\maketitle
\thispagestyle{empty}
\pagestyle{empty}

\begin{abstract}

This research explores how Large Language Models (LLMs) can assist in path planning for mobile embodied agents, such as robots, in an interactive, human-in-the-loop manner. We introduce LLM A*, a novel framework that leverages the commonsense reasoning of LLMs alongside the utility-optimal A* algorithm to enable few-shot, near-optimal path planning. By integrating LLMs within a reinforcement learning-inspired structure, LLM A* treats each interaction with the model as an alignment operation, akin to a reward function, allowing iterative refinement of the agent's policy at each step of the A* algorithm. Prompts serve two main functions: providing LLMs with essential information, such as environment details, costs, and heuristics, and relaying human feedback on intermediate planning results. This approach enhances transparency while enabling code-free path planning, fostering accessibility for users with limited coding expertise. Comparative analysis against A* and reinforcement learning (RL) shows that LLM A* achieves paths comparable to A* while significantly reducing search space and outperforming RL. Its interactive nature also makes it a promising tool for collaborative human-robot tasks. Code and supplemental materials are available on GitHub: https://github.com/speedhawk/LLM-A-.

\end{abstract}

\section{INTRODUCTION}

Path planning, alongside mapping, localisation, and motion planning, has long been regarded as the cornerstone technique for enabling fully autonomous agents such as robots~\cite{wang2015grey,zhang2019improved,wang2021feature}. It typically involves taking the environment map, the initial and goal states as inputs, and aims to find the (near) optimal path in terms of action costs to move the agent from the initial state to the goal state. Various approaches to path planning have been proposed, which can be classified into 1) search-based algorithms such as the classical A* search which is guaranteed to find an optimal path when heuristics are admissible and consistent; 2) sampling-based planning such as the Rapidly-exploring Random Tree (RRT)~\cite{gonzalez2015review} and Ant Colony Optimisation (ACO)~\cite{carr2022fast}, etc.; and 3) Data-driven planning such as Reinforcement Learning (RL) based search~\cite{zhou2022review}.

With the widespread application of neural networks, some hybrid algorithms such as neural A*~\cite{yonetani2021path} that combine traditional search algorithms with the multi-layer network cognitive model (MLP) have attracted attention from academia and industry. Although these works have greatly improved the applicability of traditional search algorithms in path planning tasks in complex environments, some of their significant shortcomings still face challenges and further research. For example, 



\begin{figure}[t]\label{fig:llmvsllmastar}
    \centering
    \subfigure[\textbf{LLM}]{\includegraphics[width=0.40\linewidth]{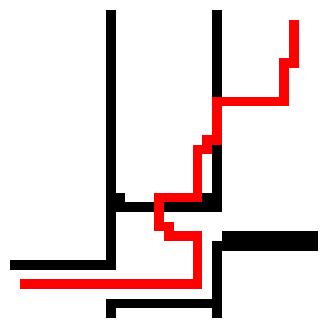}}
    \subfigure[\textbf{LLM A*}]
    {\includegraphics[width=0.40\linewidth]{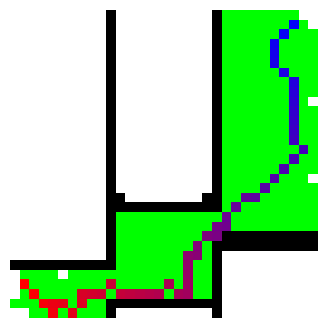}} 
    \subfigure[\textbf{Conversation between human and LLM}]
    {\includegraphics[width=0.90\linewidth]{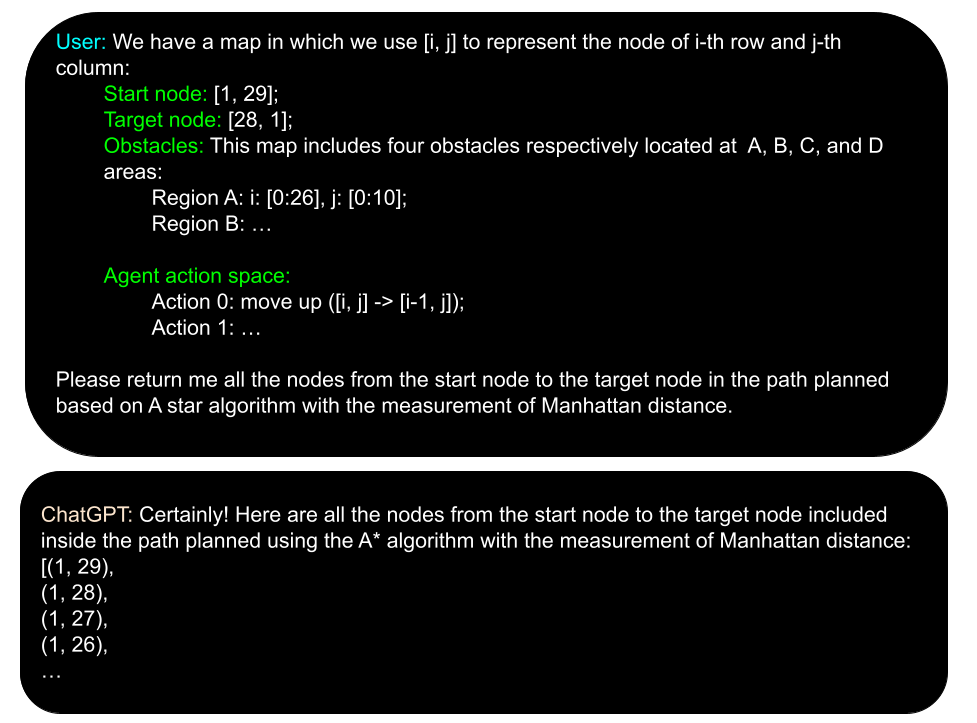}} 
    \caption{LLM-based path planning: (a) path planned by LLM directly; (b) path planned by the proposed LLM A*. The initial and goal states are at the upper right and the lower left corners, respectively. The white and black tiles represent free spaces and obstacles. The red tiles form the final paths and the green tiles are the total searched tiles. We can see the path planned by LLM goes through obstacles (results generated by GPT3.5-turbo), which is prohibitive in robotics.}
\vspace{-6mm}  
\end{figure}

Recent advancements in LLMs have facilitated a potential overhaul of data-driven techniques. LLMs exhibit capabilities akin to human-like text generation and task completion ~\cite{openai-chatgpt}. Their generated texts consistently demonstrate adept handling of commonsense information, often to a human-like level. Particularly noteworthy is their ability to engage in a chain-of-prompts interaction, wherein humans can guide LLMs to provide desired answers. This process resembles RL apart from that the rewards are assessed and granted by humans through interaction with LLMs. These developments have somewhat diverted attention from questioning the explainability of LLMs' underlying mechanisms, such as the `transformer', a type of deep neural network. Instead, the focus has shifted towards the models' proficiency in promptly responding to queries posed to them.

This paper harnesses the advantages of A* being able to find an optimal path and LLMs being able to consider commonsense knowledge when interacting with humans (agents) and proposes the LLM A*, a human-in-the-loop solution for robot path planning. 

Briefly, LLM A* can be summarised into three pivotal stages. The first stage is to set up the environment where a robot performs path planning, and communicates the information to LLMs. The second stage entails the definition of the initial state, the goal state, and the heuristics to be used. We have adopted a generic form of heuristic that considers obstacles, distances, and actions the robot can perform in the environment. It is worth mentioning that the heuristic can be tailored to specific setups based on the applications. Subsequently, LLMs identify feasible moves for path planning, evaluated against a cost function combining path cost and current-state heuristic, akin to A*. The best `move' with the minimum cost will be selected and communicated to humans upon request, who may accept or decline it. The decision will be communicated back to LLMs to carry on completing or terminating the planning. One benefit of human guidance and assessments is to minimise the use of tokens, which remains a challenge to use LLMs such as ChatGPT~\cite{openai-chatgpt}.

LLM A* adheres to a pattern of `\{state, action, rewards\}' that is similar to RL, yet without necessitating explicit coding, and humans have full control of the planning process. To this end, we have compared it to the advanced RL algorithm PPO for performance evaluation. A quantitative comparison with A* is also conducted. The contributions of this paper include: 1). a first-of-its-kind LLM-based A* is proposed for robotic path planning and a set of metrics are defined to evaluate the performance; 2). the interactive and code-free nature of LLM A* makes it appealing to non-expert robotic users; 3). enhanced safety assurance in comparison to A* and RL as humans are involved in the planning and loop and have full control of the planning process.

The remainder of the paper is organised as follows. Some related works are introduced in Section II. Section III elaborates on the approach. Experiments, discussions, and an ablation study are provided in Section IV, and the paper is concluded in Section V.

\section{Related Work}
\subsection{LLM for Robotic Task Planning}
The concept of commanding robots through human language or instructions has been a longstanding aspiration among robotcists~\cite{anderson2018vision,carr2022human}. Before the notable success of LLMs, Anderson et al.~\cite{anderson2018vision} introduced the R2R navigation framework, where Transformers were used to achieve visually grounded natural language navigation by translating human instructions into robot action sequences. Each action facilitated the robot's movement from one viewpoint to another. However, details regarding how the path was planned between viewpoints were not provided, and navigation errors of up to 10 meters were reported. 

Trained on (text) datasets of immense size, LLM models come with commonsense knowledge, which is promising to help accelerate robotic task planning~\cite{song2023llm}. To the best of our knowledge, most research from literature follows the visual-linguistic grounding strategy and proposes to harness the commonsense knowledge of LLMs to build a better correspondence between human language/instructions and visual perceptions of the robots. Song et al.~\cite{song2023llm} have demonstrated that LLMs can help with few-shot planning, i.e., the commonsense knowledge of LLMs is used to generate hierarchical plans, with the high-level plan being a set of viable while commonsense aligned subgoals generated by LLMs. The high-level plan is next passed to a low-level planner that is independent of the instructions to plan paths between subgoals. Other works such as~\cite{xie2023language} still fall into visual-linguistic grounding but a collision check is proposed.

It is worth noting that the works aforementioned do not delve into how LLMs can be employed to plan paths between subgoals or viewpoints. Aghzal et al.~\cite{aghzal2023can} introduce path planning from natural language (PPNL), emphasising the utilisation of LLMs in spatial-temporal path planning. However, their method primarily relies on prompts, lacking a connection or comparison to traditional path planning methods like A* and RL.

\subsection{Reinforcement Learning for Robotic Task Planning}
Reinforcement learning presents another data-driven learning-based framework for path planning. Within the RL framework, an agent endeavors to learn a policy that selects a sequence of actions leading it efficiently from one state to another, ultimately reaching the goal. This learning process is guided by a policy that rewards favorable actions while penalizing unfavorable ones. 

The capacity of deep learning in data representation and processing has garnered significant attention from both the RL and robotic communities. Consequently, deep learning models have been integrated into the RL framework, giving rise to deep reinforcement learning (DRL)~\cite{mnih2013playing}, to enable more sophisticated and effective policy learning, for robotic task planning, etc.~\cite{zhou2023hybrid}.

Nevertheless, both RL and DRL still face the challenges of 1) low data efficiency that necessitates a substantial number of interactions with the environment to learn policies, as well as the need for large datasets for training deep learning models; and 2) the stability of convergence, i.e., RL/DRL models are not guaranteed to converge. As a result, recent DRL models have focused on striking a balance between optimal states and convergence stability. The PPO model with actor-critic structures is one of the most prominent solutions. PPO is a policy gradient algorithm designed to maximise the horizon returns, while the actor-critic structure enables separate training of states and policy. This arrangement enhances planning performance and convergence of the PPO~\cite{andrychowicz2020matters}. 

The prompt-driven LLM-based path planning can be viewed as a variant of RL models, wherein rewards are administered by humans via prompts, and policies are implicitly learned by LLMs through interactions. Consequently, we undertake a comparison of path planning performance between LLM-based methods and RL-based methods in this paper.

\section{The LLM A* Approach} 

In most real-world scenarios, path planning is defined as the process where the goal location is known, but the environment encountered by the agent during pathfinding is uncertain. In real-world A* path planning, both distance and environmental factors must be considered. Unlike many prior studies that divide known paths into stages, we propose that pathfinding should be achieved through staged inference. In this approach, the agent dynamically selects a sub-goal, advances toward it, and continues this process until reaching the final destination.

On the other hand, the uncertainty in the environment may lead the agent to visit points that do not directly contribute to the current pathfinding task. However, these seemingly irrelevant points could hold value for future exploration tasks. Therefore, the agent should retain these "key points" for future use. After each stage of A* path planning, these points are communicated to the large language model (LLM) to assist in determining the next sub-goal.

In the following section, we present our hierarchical model, which consists of two levels: a lower level that handles A* sub-goal path planning and self-adaptive auxiliary tasks, and a higher level where the LLM determines the sub-goal based on the agent's exploration history. In the final section, we explain the interaction between these two levels.

\subsection{A* Preliminary}
In a segmented map, the cost function of A* algorithm for planning the $s$-th state is defined as: 
\begin{equation}
    f(s)=\min_{\zeta(s)\sim C}g(s) + \min_{\zeta(s)\sim C}h(s)
\end{equation}
where $g(s)$ and $h(s)$ are respectively representing the cost from the current node to the start and the goal point. For the purpose of planning, we select the minimum cost value of both $g(s)$ and $h(s)$ based on the cost policy function $\zeta(s)$, which varies as the state $s$ changes, and the cost space $C$. For the segmented map with static and known environment, obviously, the cost policy function is to calculate the distance values as the unique planning information. However, if the environment is not public to the agent at the beginning of planning, it is certain that the agent needs to gradually enrich and improve its perception of the environment in the exploration in the future. Here, we further decomposed the cost policy function into two items:
\begin{equation}
    \zeta(s) = d(s) + l(s)
\end{equation}
where $d(s)$ represents the distance value from the start and goal and $l(s)$ the relative environment value that includes the information except for distance. If we set $l(s)$ is learnable by the parameter $\theta$, the original cost function should be modified as:
\begin{equation}
    f(s)=\min_{{\zeta}_\theta(s)\sim C}g(s) + \min_{\zeta_\theta(s)\sim C}h(s)
\end{equation}
where
\begin{equation}
    \zeta_\theta(s) = d(s) + l_\theta(s)
\end{equation}
\subsection{Self-Adaption Auxiliary Task}
Enlighten by [3], our strategy to achieve the self-adaptive learning of environment value is to introduce a pixel control auxiliary task (PCAT), in which there will be a reward value that does not depend on the current state but the pixel discrepancy between two different states. Commonly, the rewards with larger values are caused by more radical changes of the pixel colour in order that the agent can 'learn more from the unprecedented experience. This reward value can be gradually learned under an semi-A2C learning structure, in which for each step, the environment value $l_\theta(s)$ of A* is equal to the fixed $V(s)$ calculated by the value network but the policy network does not need to train because that it is just the heuristic function of A* itself. Therefore, we have
\begin{equation}
    l_\theta(s) = V_\theta(s)
\end{equation}
and the value network training obeys the loss function:
\begin{equation}
    J(\theta)=\underset{sae}{\min}\frac{1}{2}(V(s)-V_\theta(s))^{2}
\end{equation}
where \begin{equation}
    V(s):=V(s)+\alpha(r(s,f(s))+\gamma V(s')-V(s))
\end{equation}
\subsection{LLM-Based Initial Reward}
It is important to note that the initial random reward $r(s,f(s))$ in deep Q-learning has not yet been fully defined. To address this issue, we introduce a large language model (LLM) as a higher-level determiner, forming a hierarchical structure in conjunction with a self-adaptive A* algorithm. At the beginning of each sub-stage of pathfinding using A*, the process from the previous stage is comprehensively evaluated and the results are fed into the LLM. This generates a data structure that informs the initial reward distribution for the current pathfinding stage.

Moreover, during A* path planning, certain jumps inevitably occur at specific points to minimize the cost function $g(s)$. In our experiments, we recorded these jump occurrences as additional feedback for the LLM. These jumps indicate that sufficient exploration has occurred in the current direction, even though the path along the jump point may not be followed further. Nevertheless, these points remain potentially valuable for future path planning tasks. By remembering these points, the LLM can more effectively determine feasible goals in subsequent stages.

\section{Experiments and Discussions}

\subsection{Setup}
To evaluate the performance of LLM A*, we conducted a series of experiments comparing it to A* and RL. We considered two variants of LLM A*:
\begin{enumerate}
    \item Greedy LLM A*, which only considers the heuristic $h(s)$ as the cost function. To avoid ambiguity, we denote it as LLM Greedy hereafter.
    \item LLM A*, which considers the combination of cumulative and heuristic costs $f(s)$ as the cost function.
\end{enumerate}
The GPT3.5-turbo-16k LLM is used as it provides more tokens than GPT3.5-turbo or even GPT4. This is to ensure that we can get the planning results without disruption, not necessarily that the algorithm needs so many tokens. The GPT-3.5-turbo-16k LLM model allows for a total of 16,384 tokens, whereas GPT-3.5-turbo has a maximum of 4,096 tokens. In case there is a need to reduce the consumption of tokens to replicate the experiments, one can reset the request dictionary to null after each interaction. To enable full control/access to the planning process, one can split a single interaction process into two stages: 1) planning by LLM A*; and 2) outputting necessary results upon requests. 

For evaluation purposes, we primarily utilise three different occupancy grid maps (which are popular in planning and navigation) denoted as Ailse, Canyon, and Double Door, respectively. Each map is of the same size $24 \times 24$ but with different obstacle distributions. Each map consists of both free spaces and obstacles, wherein the robot agent can only traverse through free spaces while avoiding collisions with obstacles. The agent can move in eight directions at most, provided it is safe to do so. All experiments are conducted at least three times using Python 3.8+ on Google Colab, and the average results are reported. The RL model employed in our experiments is based on the PPO configuration. We use this RL model alongside the standard A* algorithm for comparison purposes. Experiments on grid maps sized $16 \times 16$ and $32 \times 32$ were also conducted with results and analysis provided in the Ablation section.

\subsection{Evaluation Metrics}
To evaluate the performance of the proposed approaches and their counterparts, we utilise three metrics: \textbf{1)} The Search Complexity: For both A* and LLM A*, the overall length of the open set (grids to visit) and the closed set (visited grids) maintained by the two algorithms is used as the search complexity indicator. For LLM Greedy and RL, search complexity is equivalent to the total number of grids accessed until the goal is reached. This metric is used to explore the potential of LLM in reducing search complexity. \textbf{2)} The Path Length: Measures the length of the final path found by each algorithm. \textbf{3)} The Maximum Deviation Times (MDT): The MDT serves as a measure of the smoothness of the final path and it is defined in Equation (\ref{eq:mdn})
\begin{equation}\label{eq:mdn}
     \text{MDT}=\sum_{t=1}^{T}\delta\Big( \langle v_{d}, v_{t} \rangle\Big),
\end{equation}
where
\begin{equation}
     \delta\Big(\langle v_{d}, v_{t} \rangle\Big)=\begin{cases}
 & 1,\text{ } \pi/2<  \langle v_{d}, v_{t} \rangle\leqslant \pi \\ 
 & 0,\text{ } \text{else} 
\end{cases}.
\end{equation}
In the definitions above, $v_d$ is a directional vector that connects the initial state and the goal state and it indicates a global direction the search should expand towards. $v_t$ is a vector originating from state $s_{t-1}$ and terminating at state $s_{t}$ and we use it to represent the direction of the agent's movement from state $s_{t-1}$ to $s_{t}$. We first calculate the angle $\langle v_{d}, v_{t} \rangle$ between $v_{d}$ and $v_{t}$. If the result satisfies Equation (\ref{eq:angle}), it suggests that the agent is likely to move away from the goal. This can result in a longer path or a path with numerous back-and-forth movements, which undermines the smoothness of the path and is not preferred by agents such as robots.
\begin{equation}\label{eq:angle}
     \pi/2<  \langle v_{d}, v_{t} \rangle\leqslant \pi,
\end{equation}
where $\langle v_{d}, v_{t} \rangle$ is the inner product of $v_{d}$ and $v_{t}$, and we use it to denote the angle between $v_{d}$ and $v_{t}$. Essentially, MDT is the maximum number of times Equation (\ref{eq:angle}) is satisfied when the agent moves toward the goal along the planned path.

\begin{figure*}[ht]
    \centering
    \subfigure[]{\includegraphics[width=0.23\textwidth]{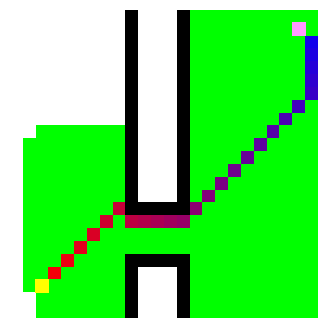}}
    \subfigure[]{\includegraphics[width=0.23\textwidth]{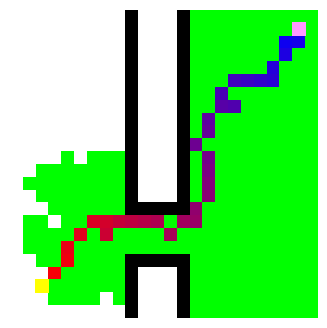}}
    \subfigure[]{\includegraphics[width=0.23\textwidth]{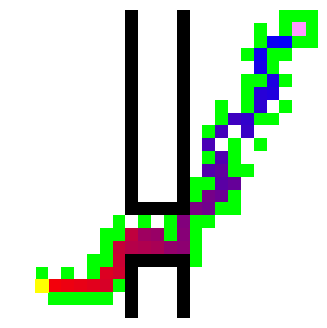}}
    \subfigure[]{\includegraphics[width=0.23\textwidth]{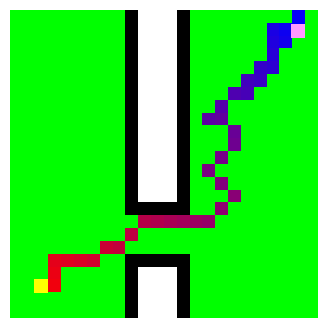}}
    \caption{Path planning results: (a), (b), (c), and (d) are results from A*, LLM A*, LLM Greedy, and PPO, respectively. The pink grids mark the initial states, while the yellow grids denote the goal states. White tiles depict free spaces, whereas black tiles represent obstacles. The green grids illustrate the search space. The final path is depicted by grids transitioning from blue to red, with color gradients aiding in visualising any back-and-forth movements within the paths.}
    \label{fig:astarRL}
\end{figure*}

\subsection{LLM A* Training and Session Design}
For experiments involving LLM, there are two stages: initialisation and interactive planning. During initialisation, essential information about the environment and the agent is prompted to be set up for planning. This includes 1) locations (coordinates) of the initial and goal states;  2) how obstacles are distributed in the environment; 3) the action space of the agent; 4) the Manhattan distance used for cumulative and heuristic cost calculation; and 5) the objective to plan a path between the initial and goal states. In addition, planning rules are communicated to LLM, such as 1) a viable path should avoid colliding with obstacles; 2) the path should ideally expand along the direction from heuristics or human guidance; 3) preference of actions that help accelerate the planning process. 

In the second interactive planning stage, LLM can provide planning results based on the initial information and other prompts to assist humans in guiding or monitoring the planning process. This stage operates iteratively and interactively until a path is successfully planned. It's important to note that humans can request intermediate planning results at any stage, making the planning process transparent (white box) to humans and ensuring safety, among other benefits.

\subsection{RL Model Training}
The PPO model serves as a data-driven benchmark alongside A* for comparison. In our setup, the PPO model adopts an actor-critic structure comprising two 3-layer deep neural networks for policy and value training, respectively. The model undergoes training for 2,000 episodes, with each episode having a maximum of 200 steps to ensure convergence. To prevent premature model fixation and achieve a balance between exploration and exploitation, we set the learning rates of both the actor and critic to be smaller than 0.0005.

\begin{figure}[hp]
    \centering
    \subfigure[]{\includegraphics[width=0.23\textwidth]{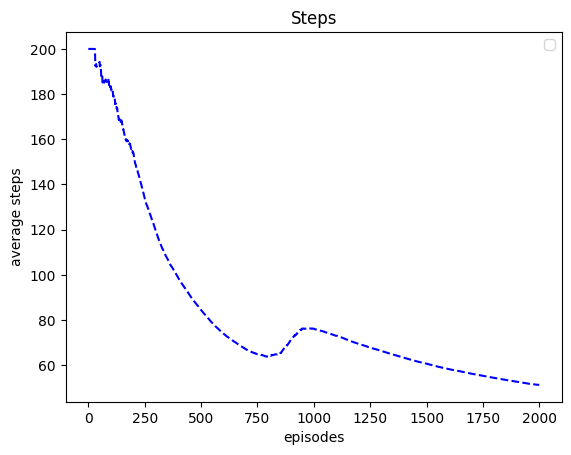}}
    \subfigure[]{\includegraphics[width=0.23\textwidth]{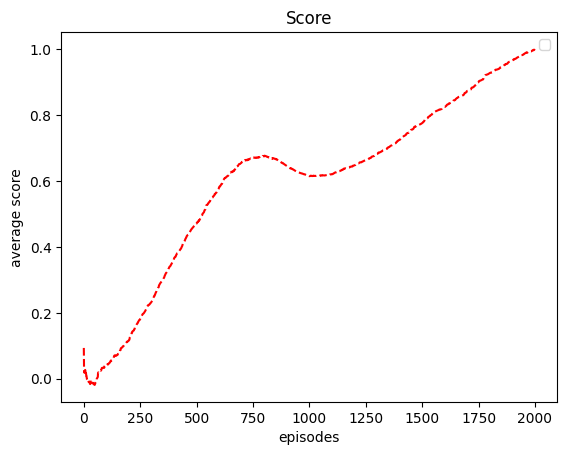}}
    \caption{Covergence of RL model training: (a) average steps per episode; (b) average scores achieved per episode.}
    \label{fig:RLconverge}
\end{figure}

To enhance the adaptability of the PPO model to diverse environments, we introduce a mechanism of randomising the initial state in each episode. This mechanism involves a progressive training approach, starting from states near the goal where the agent is close enough to reach it in one step. Subsequently, states are randomly selected from this subset for PPO training. As the model initially operates within a limited map scope, convergence is relatively easy. We then incrementally increase the map scope during subsequent training iterations until convergence is achieved for path planning from the initial state to the goal state. This process, known as the easy-to-difficult mechanism, enables the model to balance adaptability to varying environments while efficiently converging. Fig. \ref{fig:RLconverge} shows the average steps and scores against episodes for the PPO model training. We can see that the PPO model converges in 2,000 episodes and the results are eventually generated from this model.


\begin{figure*}[tb]
    \centering
    \includegraphics[width=0.95\textwidth]{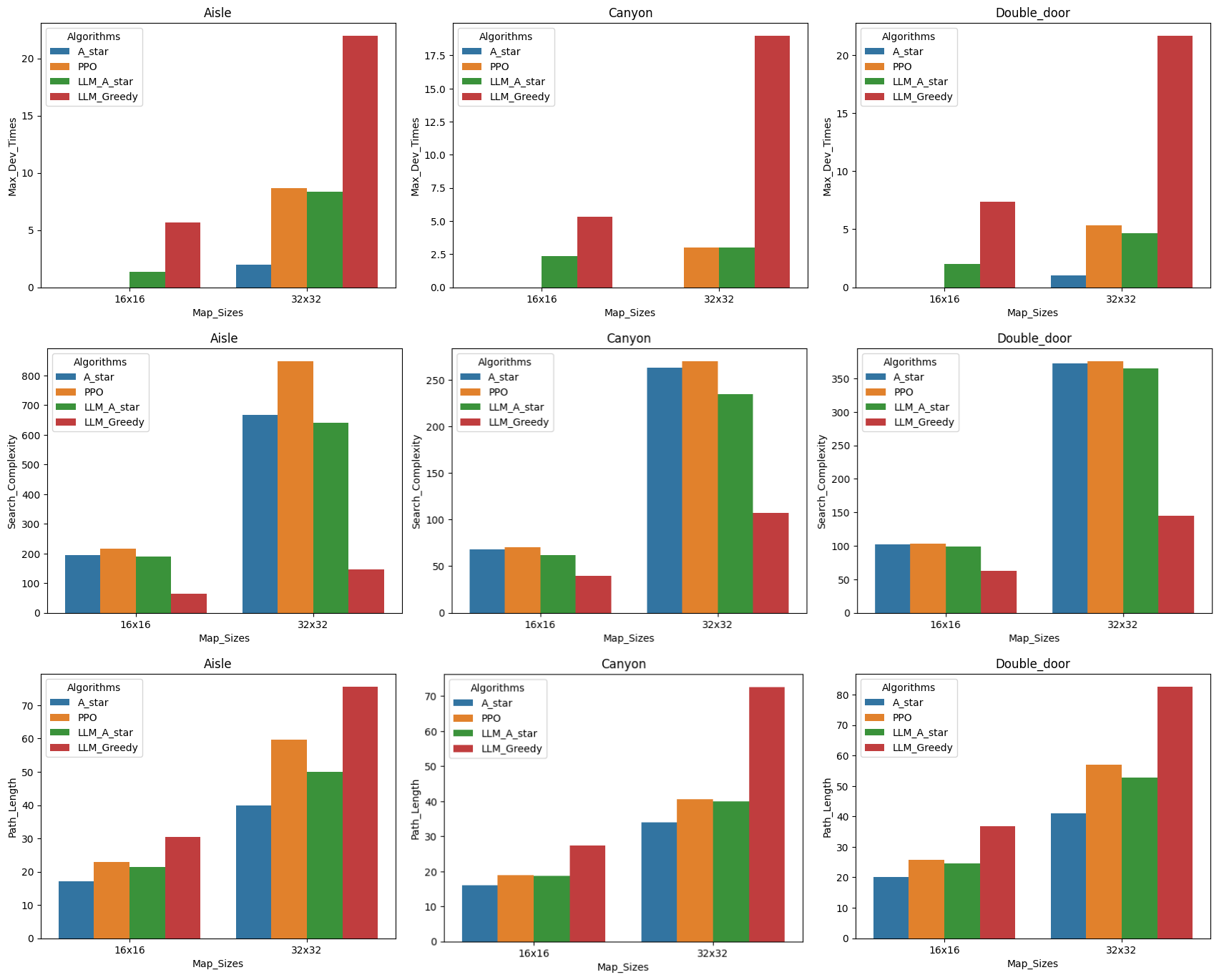}
    \caption{Experimental results of A*, LLM A*, LLM Greedy, and PPO on the Aisle and Double Door environments with different sizes, i.e., $16\times16$ and $32\times32$.}
    \label{fig:Canyon_24x24}
\end{figure*}

\subsection{Main Results}
Figure \ref{fig:astarRL} illustrates the visual comparison of path planning results obtained from A*, LLM A*, LLM Greedy, and the PPO model. It is evident that the search complexity, indicated by the green tiles, of LLM A* is reduced compared to A* and PPO. However, it is notable that LLM Greedy outperforms LLM A*. This is attributed to the greedy nature of the heuristics, which consistently guides the agent towards the goal direction. However, the presence of obstacles may cause the agent to be redirected backwards, resulting in back-and-forth movements in the final path, making it less efficient. On the other hand, LLM A* considers both cumulative and heuristic costs, leading to an expanded search space but resulting in a smoother path with fewer back-and-forth movements.

Table \ref{tab:table24by24} shows the quantitative results achieved by each algorithm against Path Length, MDT, and Search Complexity. Notably, LLM-based algorithms demonstrate superior performance compared to PPO in all environments and maintain an advantage over A* in most cases in terms of search complexity. In addition, LLM A* outperforms PPO and LLM Greedy in terms of path length, achieving comparable results with A*. However, A* exhibits superiority in terms of path smoothness, with LLM A* and PPO showing similar performance. Further studies have been carried out to further investigate and evaluate the performance in various environments, and the results are given in the Ablation section and the Supplemental Materials in our Github Repo: https://github.com/speedhawk/LLM-A-.

\begin{table}[t!]
    \centering
    \begin{tabular}{l|l|llll}
    \hline
     Metrics    &  Environments &  A* &  LLM A* &  LLM Greedy & PPO\\
     \hline
         &  Aisle &  \textbf{27}&  \textbf{34}&  48.67& 37.33\\
         Path Length&  Canyon &  \textbf{24}&  \textbf{27.67}&  41.67& 29\\
         &  Double Door &  \textbf{31}&  \textbf{37.33}&  45.67& 38\\
         \hline
         &  Aisle &  \textbf{1}&  4.67&  12.33& \textbf{3}\\
         MDT &  Canyon &  \textbf{0}&  2.33&  7.67& \textbf{0.33}\\
         &  Double Door &  \textbf{0}&  3&  7& \textbf{2.33}\\
         \hline
         &  Aisle &  372&  \textbf{352}&  \textbf{100.67}& 471\\
         Complexity &  Canyon &  145 &  \textbf{133 }&  \textbf{74 }& 156\\
         &  Double Door &  210 &  \textbf{197.67}&  \textbf{103.67}& 214\\
         \hline
    \end{tabular}
    \caption{Evaluation results of the algorithms against the defined metrics. Note each algorithm was executed three times and the average results are reported. The best two results are highlighted in the table.}
    \label{tab:table24by24}
\end{table}

\subsection{Ablation Study}
For a more comprehensive analysis of performance, additional experiments were conducted on grid maps sized $16 \times 16$ and $32 \times 32$. The results are depicted in Figure \ref{fig:Canyon_24x24}. It is evident that LLM A* generally outperforms both LLM Greedy and PPO across all metrics in each environment while achieving comparable results with A*.

Notably, there is an increasing trend in MDT for all algorithms as the size of the environment increases. However, we have observed a comparative advantage of LLM A* over PPO as the environment size increases, indicating that the agent following the path planned by LLM A* is less likely to experience back-and-forth movements within the final path compared to the path planned by RL. This underscores the potential of human knowledge and commonsense from LLMs to enhance path-planning outcomes in robotics.

\subsection{Discussion}

The integration of LLMs undoubtedly leads to increased interactions between humans and AI agents. Our research has shown that by incorporating LLMs into traditional path-planning algorithms like A*, we can achieve near-optimal path-planning results while still allowing humans to oversee the process. This represents a significant step towards making AI techniques transparent to humans and contributes to ensuring safety in human-robot interaction and collaboration.

It is important to highlight that the process of planning a path through interaction with LLMs bears a resemblance to using RL in path planning. Our findings demonstrate that leveraging commonsense knowledge from LLMs can greatly enhance path planning performance, particularly in terms of path length, path smoothness, and search complexity. This aspect will be further explored in our future work.

\section{CONCLUSIONS}

This paper makes the first effort with the integration of Large Language Models (LLMs) with the classic A* algorithm, introducing LLM A* as a novel approach to human-in-the-loop interactive path planning for mobile embodied agents. By leveraging the inherent commonsense of LLMs and the optimality of A*, LLM A* achieves few-shot near-optimal path planning compared to data-driven models such as PPO. Additionally, LLM A* offers the unique advantage of providing humans with complete access to the path-planning process, thereby enhancing safety when incorporating LLM A* in embodied agent path planning.

Despite the notable advantages of LLM A*, it is acknowledged that it shares similar inefficiencies (the LLM model training is data and computational resources demanding) with data-driven methods such as RL models. Future research will focus on enhancing the efficiency of LLM from both language modeling training/fine tuning and its  integration with A*, aiming to strike a balance between interactive capacity and efficiency to facilitate the real-life deployment of such models.

\addtolength{\textheight}{-12cm}   







\bibliographystyle{IEEEtran}
\bibliography{IROS2025LLMASTAR}

\end{document}